\documentclass[10pt,twocolumn,letterpaper]{article}

\usepackage{cvpr}
\usepackage{times}
\usepackage{epsfig}
\usepackage{graphicx}
\usepackage{amsmath}
\usepackage{amssymb}

\usepackage{epsfig}
\usepackage{graphicx}
\usepackage{amsmath}
\usepackage{pifont}
\usepackage{amssymb}
\usepackage{multirow}
\usepackage{tabularx}
\usepackage{diagbox}
\usepackage{booktabs}
\usepackage[mathscr]{euscript}
\usepackage{blindtext}
\usepackage{booktabs}
\usepackage{amssymb,mathrsfs,amsmath}

\usepackage[normalem]{ulem}

\newcommand*{\affaddr}[1]{#1} 
\newcommand*{\affmark}[1][*]{\textsuperscript{#1}}
\newcommand*{\email}[1]{\texttt{#1}}

\usepackage[pagebackref=true,breaklinks=true,letterpaper=true,colorlinks,bookmarks=false]{hyperref}

\cvprfinalcopy 

\def\cvprPaperID{1550} 

\ifcvprfinal\fi
\begin{document}

\title{Human De-occlusion: Invisible Perception and Recovery for Humans}

\author{
  Qiang Zhou\affmark[1]\thanks{The work was mainly done during an internship at ByteDance Inc.}, \hspace{0.1cm}
  Shiyin Wang\affmark[2], \hspace{0.1cm}
  Yitong Wang\affmark[2], \hspace{0.1cm}
  Zilong Huang\affmark[1], \hspace{0.1cm}
  Xinggang Wang\affmark[1]\thanks{Corresponding author.}\\

	\affaddr{\affmark[1]School of EIC, Huazhong University of Science and Technology} \hspace{0.2cm}
	\affaddr{\affmark[2]ByteDance Inc.}\\
  \email{\tt\small{theodoruszq@gmail.com}} \hspace{0.1cm}
  \email{\tt\small{shiyinwang.ai@bytedance.com}} \hspace{0.1cm}
  \email{\tt\small{wangyitong@pku.edu.cn}} \\
  \email{\tt\small{zilong.huang2020@gmail.com}} \hspace{0.1cm}
  \email{\tt\small {xgwang@hust.edu.cn}}\\
}

\maketitle

\begin{abstract}

In this paper, we tackle the problem of human de-occlusion which reasons about occluded segmentation masks and invisible appearance content of humans. In particular, a two-stage framework is proposed to estimate the invisible portions and recover the content inside.
For the stage of mask completion, a stacked network structure is devised to refine inaccurate masks from a general instance segmentation model and predict integrated masks simultaneously. Additionally, the guidance from human parsing and typical pose masks are leveraged to bring prior information. 
For the stage of content recovery, a novel parsing guided attention module is applied to isolate body parts and capture context information across multiple scales. 
Besides, an Amodal Human Perception dataset (AHP) is collected to settle the task of human de-occlusion. AHP has advantages of providing annotations from real-world scenes and the number of humans is comparatively larger than other amodal perception datasets.
Based on this dataset, experiments demonstrate that our method performs over the state-of-the-art techniques in both tasks of mask completion and content recovery.
Our AHP dataset is available at \url{https://sydney0zq.github.io/ahp/}.
\end{abstract}

\section{Introduction} \label{introduction}

Visual recognition tasks have witnessed significant advances driven by deep learning, such as classification~\cite{imagenet_nips12,resnet_cvpr16}, detection~\cite{fastercnn_nips15,rcnn_cvpr14} and segmentation~\cite{fcn_cvpr15,dilation_arxiv15}. 
Despite the achieved progress, \textit{amodal perception}, \ie to reason about the occluded parts of objects, is still challenging for vision models. 
In contrast, it is easy for humans to interpolate the occluded portions with human visual systems~\cite{cocoa_cvpr17, photo2phen_MIT99}.
To reduce the recognition ability gap between the models and humans, recent works have proposed methods to infer the occluded parts of objects, including estimating the invisible segmentation~\cite{amodalseg_eccv16, cocoa_cvpr17, kins_cvpr19, caramodal_iccv19, pcnets_cvpr20} and recovering the invisible content of objects~\cite{segan_cvpr18, caramodal_iccv19, pcnets_cvpr20}.

\begin{figure}[t]
  \centering
  \includegraphics[width=1.0\linewidth]{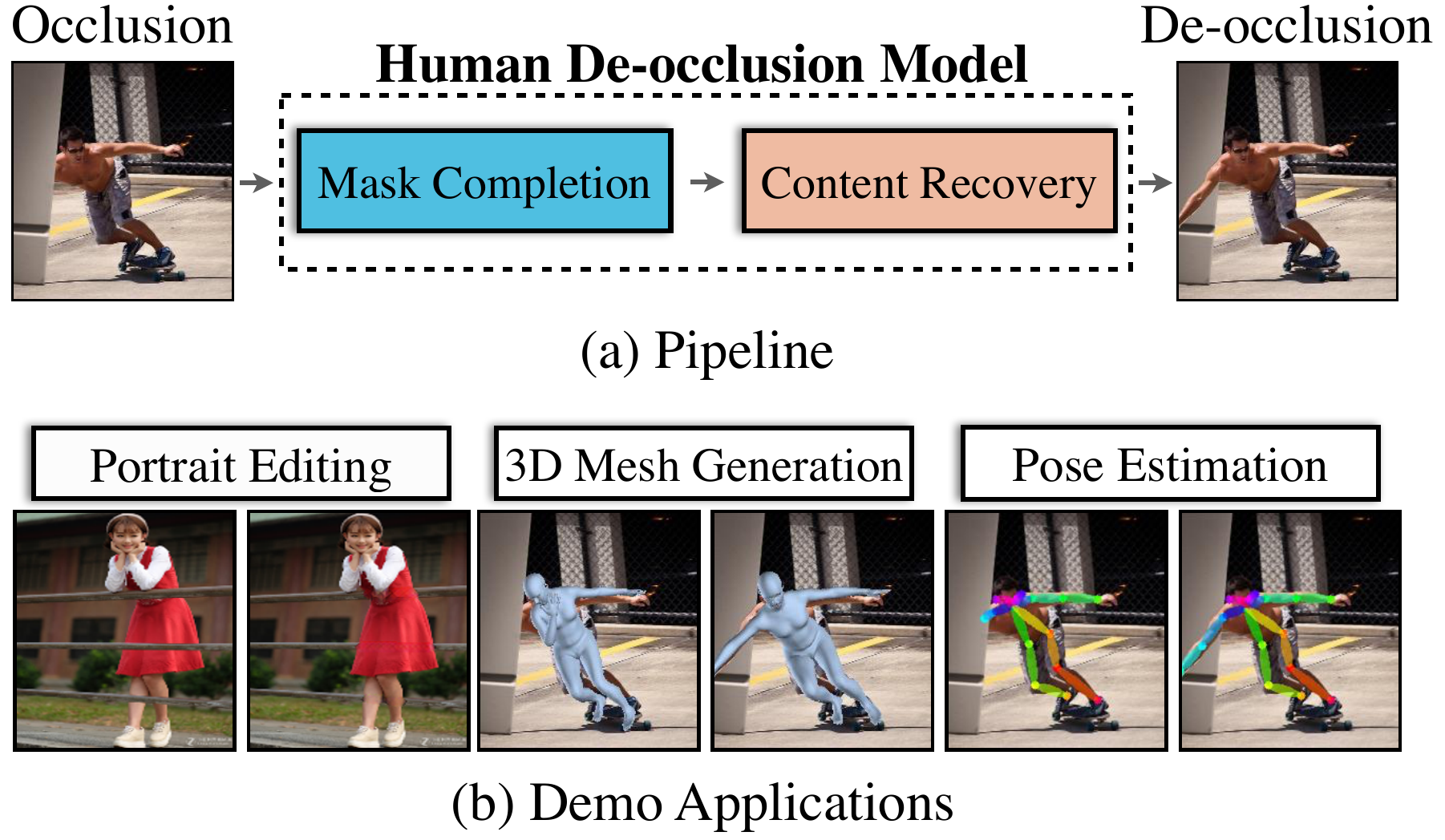}
  \caption{(a) The pipeline of our framework to tackle the task of human de-occlusion, which contains two stages of mask completion and content recovery. (b) Some applications which demonstrate better results can be obtained after human de-occlusion.}
  \label{fig:introduction}
\end{figure}

In this paper, we aim at the problem of estimating the invisible masks and the appearance content for humans. We refer to such a task as \textit{human de-occlusion}. Compared with general amodal perception, human de-occlusion is a more special and important task, since completing human body plays key roles in many vision tasks, such as portrait editing~\cite{imageinpainting_bertalmio_cgit2000}, 3D mesh generation~\cite{hmr_cvpr18}, and pose estimation~\cite{lip_cvpr17} as illustrated in Fig~\ref{fig:introduction} (b). Different from common object de-occlusion task (\eg, vehicle, building, furniture), human de-occlusion presents new challenges in three aspects. First, humans are non-rigid bodies and their poses vary dramatically. Second, the context relations between humans and backgrounds are relatively inferior to common objects which are usually limited to specific scenes. Third, to segment and recover the invisible portions of humans, the algorithm should be able to recognize the occluded body parts to be aware of symmetrical or interrelated patches.

To precisely generate the invisible masks and the appearance content of humans, we propose a two-stage framework to accomplish human de-occlusion as shown in Fig~\ref{fig:introduction} (a). The first stage segments the invisible portions of humans, and the second stage recovers the content inside the regions obtained in the previous stage. In the testing phase, the two stages are cascaded into an end-to-end manner.

Previous methods~\cite{pcnets_cvpr20, segan_cvpr18, caramodal_iccv19} adopt perfect modal~\footnote{The terms `modal' and `amodal' are used to refer to the visible and the integrated portions of an object respectively.} mask as an extra input (\eg, ground-truth annotation or well-segmented mask), and expect to output amodal mask. However, obtaining ground-truth or accurate segmented masks is non-trivial in actual applications.
To address the issue, our first stage utilizes a stacked hourglass network~\cite{hg_eccv16} to segment the invisible regions progressively. Specifically, the network first applies a hourglass module to refine the inaccurate input modal mask, then a secondary hourglass module is applied to estimate the integrated amodal mask. In addition, our network benefits from human parsing and typical pose masks. Inspired by ~\cite{shapemask_iccv19, caramodal_iccv19}, human parsing pseudo labels are served as auxiliary supervisions to bring prior cues and typical poses are introduced as references.

In the second stage, our model recovers the appearance content inside the invisible portions. Different from typical inpainting methods~\cite{generalinpainting_cvpr16, partialconv_eccv18, gatedconv_iccv19, edgeconnect_iccvw19, lbam_iccv19}, the context information inside humans should be sufficiently explored. To this end, a novel parsing guided attention (PGA) module is proposed to capture and consolidate the context information from the visible portions to recover the missing content across multiple scales. Briefly, the module contains two attention streams. The first path isolates different body parts to obtain relations of the missing parts with other parts. The second spatially calculates the relationship between the invisible and the visible portions to capture contextual information. Then the two streams are fused by concatenation and the module outputs an enhanced deep feature. The module works at different scales for stronger recovery capability and better performance.

As most existing amodal perception datasets~\cite{cocoa_cvpr17, kins_cvpr19, sailvos_cvpr19} focus on amodal segmentation and few human images are involved, an amodal perception dataset specified for human category is required to evaluate our method.
On account of this, we introduce an Amodal Human Perception dataset, namely AHP. There are three main advantages of our dataset:

\textbf{a)} the number of humans in AHP is comparatively larger than other amodal perception datasets; 
\textbf{b)} the occlusion cases synthesized from AHP own amodal segmentation and appearance content ground-truths from real-world scenes, hence subjective judgements and consistency of the invisible portions from individual annotators are unnecessary; 
\textbf{c)} the occlusion cases with expected occlusion distribution can be readily obtained. Existing amodal perception datasets have fixed occlusion distributions but some scenarios may have gaps with them. 

Our contributions are summarized as follows:
\begin{itemize}
  \item We propose a two-stage framework for precisely generating the invisible parts of humans. To the best of our knowledge, this is the first study which both considers human mask completion and content recovery.
  \item A stacked network structure is devised to do mask completion progressively, which specifically refines inaccurate modal masks. Additionally, human parsing and typical poses are introduced to bring prior cues.
  \item We propose a novel parsing guided attention (PGA) module to capture body parts guidance and context information across multiple scales.
  \item A dataset, namely AHP, is collected to settle the task of human de-occlusion.
\end{itemize}

\section{Related Work} \label{related_work}

\begin{figure*}[htbp]
  \centering
  \includegraphics[width=0.9\linewidth]{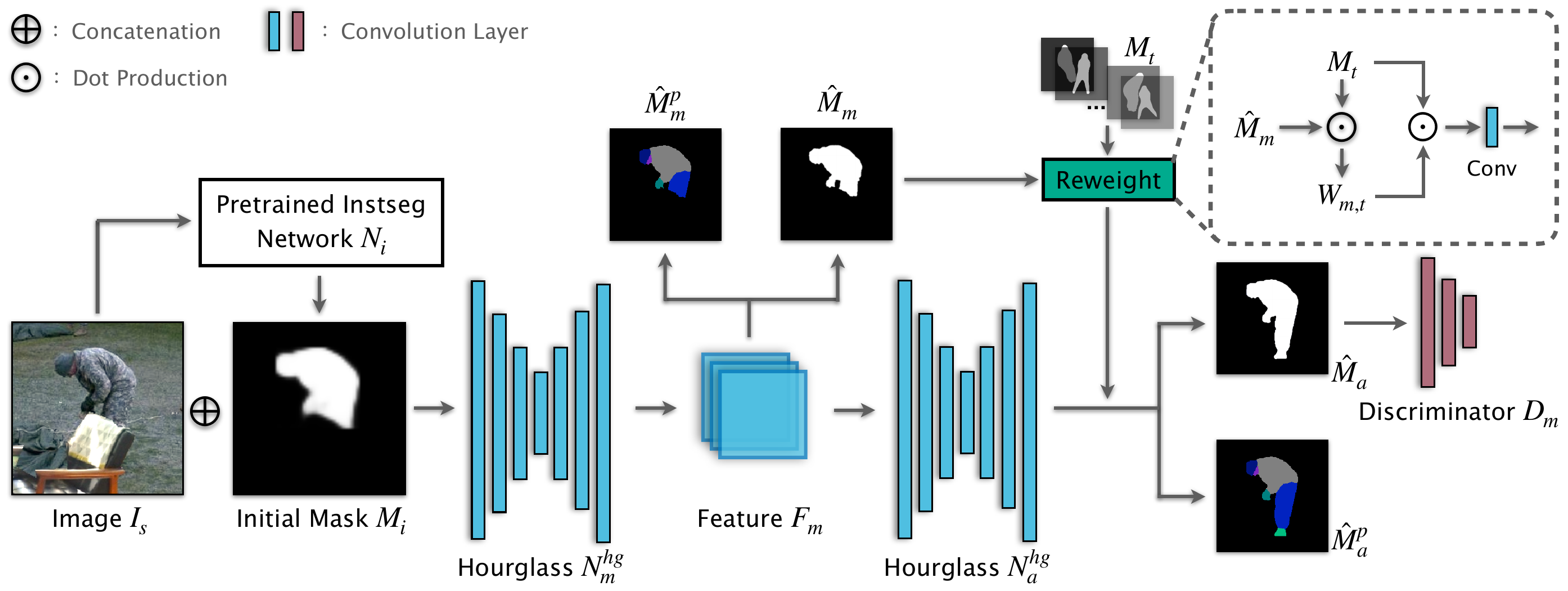}
  \caption{\textbf{Illustration of the refined mask completion network.} A pretrained instance segmentation network $N_i$ is applied to obtain the initial modal mask $M_i$ from the input image $I_s$ which contains an occluded human. Then one hourglass module $N_{m}^{hg}$ outputs a refined modal mask $\hat{M}_{m}$ and the corresponding parsing result $\hat{M}_m^{p}$. To complete the modal mask, another hourglass module $N_{a}^{hg}$ is stacked with the template masks which served as prior appearance cues and it finally outputs the amodal mask $\hat{M}_{a}$ and the parsing result $\hat{M}_p^{a}$. At the end, a discriminator $D_m$ is applied to improve the quality of the generated amodal mask.}
  \label{fig:maskcomp_net}
\end{figure*}

\noindent\textbf {Amodal Segmentation and De-occlusion.} The task of modal segmentation is to assign categorical or instance label for each pixel in an image, including semantic segmentation~\cite{fcn_cvpr15, deeplab_pami17, pspnet_cvpr17, deconv_cvpr15} and instance segmentation~\cite{fastercnn_nips15, maskrcnn_iccv17, blendmask_cvpr20}. Differently, amodal segmentation aims at segmenting the visible and estimating the invisible regions, which is equivalent to the amodal mask, for each instance. Research on amodal segmentation emerges from ~\cite{amodalseg_eccv16} which iteratively enlarges detected box and recomputes heatmap of each instance based on modally annotated data. Afterwards, AmodalMask~\cite{cocoa_cvpr17}, MLC~\cite{kins_cvpr19}, ORCNN~\cite{learn2seeinv_wacv19} and PCNets~\cite{pcnets_cvpr20} push forward the field of amodal segmentation by releasing datasets or techniques. 

De-occlusion~\cite{pcnets_cvpr20, segan_cvpr18, caramodal_iccv19} targets at predicting the invisible content of instances. It is different from general inpainting~\cite{partialconv_eccv18, gatedconv_iccv19, edgeconnect_iccvw19, lbam_iccv19} which recovers the missing areas (manual holes) to make the recovered images look reasonable. At present, de-occlusion usually depends on the invisible masks predicted by amodal segmentation. SeGAN~\cite{segan_cvpr18} studies amodal segmentation and de-occlusion of indoor objects with synthetic data. Yan \etal~\cite{caramodal_iccv19} propose an iterative framework to complete cars. Xu \etal~\cite{portrait_comp_tip19} aim at portrait completion without involving amodal segmentation. Compared with them, we jointly tackle the tasks of human amodal segmentation and de-occlusion. Not only that, humans commonly have larger variations in shapes and texture details than rigid vehicles and furniture.

\noindent\textbf {Amodal Perception Datasets.} Amodal perception is a challenging task~\cite{learn2seeinv_wacv19, kins_cvpr19, cocoa_cvpr17, segan_cvpr18} aiming at recognizing occluded parts of instances. Zhu \etal~\cite{cocoa_cvpr17} point out that humans are able to predict the occluded regions with high degrees of consistency though the task is ill-posed. There are some pioneers building amodal segmentation datasets. COCOA~\cite{cocoa_cvpr17} elaborately annotates modal and amodal masks of $5,000$ images in total which originates from COCO~\cite{coco_eccv14}. To alleviate the problem that the data scale of COCOA is too small, Qi \etal~\cite{kins_cvpr19} establish KINS with $14,991$ images for two main categories of `people' and `vehicle'. In \cite{segan_cvpr18}, Ehsani \etal introduce a synthetic indoor dataset namely DYCE. SAIL-VOS~\cite{sailvos_cvpr19} is also synthetic but a video dataset by leveraging a game simulator. Our dataset AHP focuses on amodal perception of humans on account of its wide applications. Particularly, the differences with the above datasets mainly lie in two aspects: 
\textbf{a)} our dataset greatly extends the quantity ($\sim56$k) of humans in real-world scenes and provides amodal mask annotations; 
\textbf{b)} our dataset emphasizes on completing both invisible masks and appearance content with trustworthy supervisions provided, to overcome the previous learning techniques with only modal supervisions in \cite{pcnets_cvpr20, learn2seeinv_wacv19}.

\noindent\textbf {Human Related Works.} There are extensive literature on humans, such as detection~\cite{repulsion_cvpr18, breg_eccv18, oacnn_eccv18}, parsing~\cite{hrnetocr_arxiv19, corrparsing_cvpr20, neuralinfoparsing_iccv19, singlemultiparsing_aaai19} and pose estimation~\cite{hourglass_pose_eccv16, rtpose_cvpr17, lstmpose_cvpr18, combpose_cvpr20} \etc. Our work attempts to benefit from these studies for better performance.

%
\section{Method}        \label{method}

\subsection{Overview}

\begin{figure*}[htbp]
  \centering
  \includegraphics[width=0.9\linewidth]{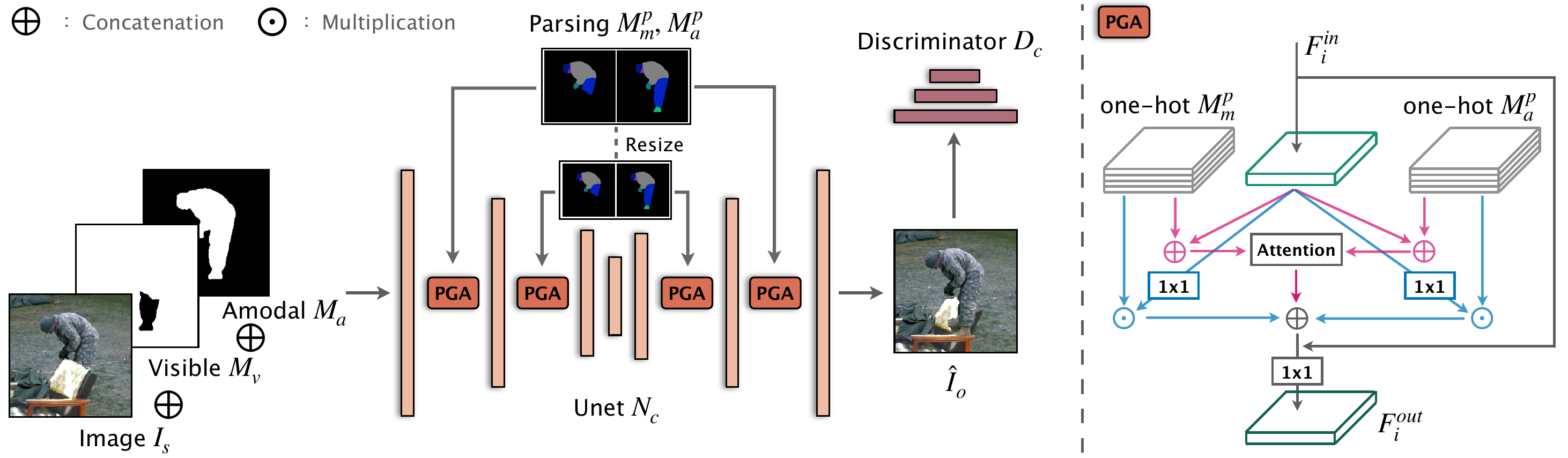}
  \caption{\textbf{Illustration of the parsing guided content recovery network.} \textbf{Left}: We adopt Unet~\cite{unet_miccai15} with partial convolution~\cite{partialconv_eccv18} as the basic architecture. The image $I_s$ concatenated with the visible mask $M_v$ and the amodal mask $M_a$ is passed into the network $N_c$ to recover content of the invisible portions. To leverage the guidance of body part cues $M_m^p$ and $M_a^p$ from the previous stage, our proposed Parsing Guided Attention (PGA) module enhances deep features at multiple scales. Finally a discriminator $D_c$ is applied to identify the quality of the output image $I_o$. \textbf{Right}: The details of our proposed PGA module which contains two attention streams to capture context relations. The first stream (\textcolor{cyan}{cyan}) decomposes the feature into different body parts and the second (\textcolor{magenta}{magenta}) tries to establish the pixel-level relationship between the visible context and the invisible regions.}
  \label{fig:rgbcomp_net}
\end{figure*}

We propose a two-stage framework to accomplish human de-occlusion. The first stage segments the invisible portions of humans and the second recovers the appearance content inside, as shown in Fig~\ref{fig:maskcomp_net} and Fig~\ref{fig:rgbcomp_net} respectively.

\subsection{Refined Mask Completion Network} \label{maskcomp_net}

The refined mask completion network targets at segmenting the invisible masks. As mentioned in Sec.~\ref{introduction}, most methods first take perfect but non-trivial modal masks as input and the predicted amodal masks will be subtracted with them to get the expected invisible masks. Some of them also apply modal masks from instance segmentation methods, but the gap between amodal perception and instance segmentation annotations are not being noticed. To this end, our mask completion network firstly refines trustless initial modal masks. Specifically, an existing instance segmentation network $N_i$ is applied to obtain the initial modal masks $M_{i}$ from the input image $I_s$ which contains an occluded human. Then the image and the initial mask are concatenated and fed into the first hourglass module $N_{m}^{hg}$ to obtain a refined binary modal mask $\hat{M}_{m}$. The supervision on corresponding parsing result $\hat{M}_m^{p}$ is accompanied to strengthen the semantic understanding of body parts. The modal recognition process can be formulated as:
\begin{equation}
  \label{eq:modal_mask}
  \hat{M}_{m},\: \hat{M}_m^{p} =\: N_{m}^{hg}\: (I_s,\: N_i\: (I_s)).
\end{equation}

Next, our network needs to complete the refined modal mask. One direct solution is to augment another amodal branch to accomplish it. However, we empirically find the direct solution will significantly degrade the performance of the refined modal mask. We suspect that completing the amodal mask requires analogous feature both in the occluded and the visible portions of the human, which is opposite to the previous modal mask recognition paying attention on the visible regions only. Accordingly, another hourglass module $N_{a}^{hg}$ is stacked behind the first one to estimate the amodal mask $\hat{M}_{a}$ and the parsing result $\hat{M}_a^{p}$.

Since it can be asserted that the target amodal masks are integrated humans, some typical poses can be implanted into the network as prior appearance cues. We collect a batch of template masks by running k-means on the annotations of the training set, denoted as $M_t$. Then the $\ell_2$ distances $D_{m,t}$ of $\hat{M}_{m}$ with each template mask is calculated and the attention weight vector can be obtained by $W_{m,t}=1/D_{m,t}$. The vector $W_{m,t}$ is multiplied back with the template masks to highlight suitable candidates and a convolution layer without activation is applied to combine these re-weighted templates as an additional feature before making predictions. The amodal completion process is given by:
\begin{equation}
  \label{eq:amodal_mask}
  \hat{M}_{a},\: \hat{M}_a^{p} = N_{a}^{hg}\: (F_m\oplus \hat{M}_{m},\: \mathrm{Conv.}(M_t \odot W_{m,t})),\\
\end{equation}
where $F_m$ is the intermediate feature from the last hourglass module.

The cross-entropy loss $\mathcal{L}_{CE}(*)$ is applied to supervise the modal and the amodal predictions and the human parsing. The idea in adversarial learning can be borrowed to boost the performance by adopting a discriminator $D_m$ and the perceptual loss~\cite{precloss_cvpr16} $\mathcal{L}_{prec}(*)$ which measures the distances of the extracted features of the generated masks and the ground-truth masks. The several loss functions are formulated as follows:
\begin{equation}
  \label{eq:loss_submask}
  \begin{aligned}
    \mathcal{L}_{seg}\: =\: &\mathcal{L}_{CE}(\hat{M}_{m}, M_{m}) + \mathcal{L}_{CE}(\hat{M}_{a}, M_{a}) + \\
                      &\mathcal{L}_{CE}(\hat{M}_{m}^{p}, M_{m}^p) + \mathcal{L}_{CE}(\hat{M}_{a}^{p}, M_{a}^p),\\
    \mathcal{L}_{adv}\: =\: &\mathbb{E}_{\hat{M}_{a}}[\mathrm{log}(1-D_m(\hat{M}_{a}))] + \mathbb{E}_{M_{a}}[\mathrm{log}\:D_m(M_{a})],\\
    \mathcal{L}_{gen}\: =\: &\mathcal{L}_{\ell 1}(\hat{M}_{a}, M_{a}) + \mathcal{L}_{prec}(\hat{M}_{a}, M_{a}),
  \end{aligned}
\end{equation}
where $\mathcal{L}_{\ell 1}(*)$ denotes the reconstruction loss. Then the final loss can be summarized with proper coefficients:
\begin{equation}
  \label{eq:loss_mask}
  \begin{aligned}
    \mathcal{L}_{m}\: = \lambda_1 \mathcal{L}_{seg} + \lambda_2 \mathcal{L}_{adv} + \lambda_3 \mathcal{L}_{gen}.
  \end{aligned}
\end{equation}

\subsection{Parsing Guided Content Recovery Network}

The mask completion network is able to localize the invisible regions and the body parts by subtracting the refined modal results from the amodal ones, and the goal of this stage is to recover the appearance content inside. Different from general inpainting methods, the missing content is some parts of a human instead of other surroundings to make the recovered images look plausible. In spite of this distinction, some well studied methods can be referred and we adopt Unet~\cite{unet_miccai15} with partial convolution~\cite{partialconv_eccv18} as our architecture, as shown on the left of Fig.~\ref{fig:rgbcomp_net}. 

Our pipeline is straightforward. At first, the visible mask $M_v$ and the amodal mask $M_a$ are concatenated with the image $I_s$ containing an occluded human as input. The visible mask tells the network $N_c$ which pixels need to be recovered and the amodal mask points out where the integrated human occupies in the image. Additionally, the body part cues $M_m^p$ and $M_a^p$ from the previous stage are aggregated into the deep features by our proposed Parsing Guided Attention (PGA) module at multiple scales. Finally the network outputs a human recovered image $\hat{I}_o$ and a discriminator $D_c$ is applied to estimate the quality of the image $\hat{I}_o$. This stage can be formulated as:
\begin{equation}
  \label{eq:rgbcomp_net}
  \hat{I}_o\: =\: N_{c}\:(I_s\oplus M_v\oplus M_a,\: M_m^p,\: M_a^p).
\end{equation}

The PGA module is depicted on the right of Fig~\ref{fig:rgbcomp_net}. Take some scale $i$ for example, the module takes in deep feature $F_i^{in}$ and the parsing masks $M_m^p$ and $M_a^p$ which are converted into two one-hot logits and resized to the same spatial size with $F_i^{in}$. It contains two attention streams.
The first stream (\textcolor{cyan}{cyan}) decomposes the feature into different body parts and compare them. Specifically, the feature $F_i^{in}$ is reduced to the same channel number with the parsing logits (\ie $19$), and it is element-wisely multiplied with the two logits to distribute the feature in different body parts. Then the two distributed features are concatenated and a $1\times 1$ convolution layer is applied to yield a feature of useful body parts.
The second stream (\textcolor{magenta}{magenta}) tries to establish the pixel-level relationship between the visible context and the invisible regions. The difference with the self attention~\cite{nonlocal_iccv18} is that self attention wildly calculates pixel-wise relations regardless of the semantics of pixels. In particular, we concatenate the two logits behind the input feature, and two $1\times1$ convolution layers $\phi(\ast)$ and $\psi(\ast)$ are followed to extract key features for the visible and the amodal regions respectively, denoted as $K_{vis}=\phi(F_i^{in}\oplus M_m^{p})$ and $K_{amo}=\psi(F_i^{in}\oplus M_a^{p})$. Then a relationship matrix $R$ can be obtained by:
\begin{equation}
  \label{eq:attention}
  \begin{aligned}
  \tilde{R} &= (M_v\odot K_{vis})^T\:((1-M_v)\odot K_{amo});\\
  R &= \mathrm{Softmax}(\tilde{R},\: \mathrm{dim}=0) \in \mathbb{R}^{HW\times HW},\\
  \end{aligned}
\end{equation}
where $H$ and $W$ denote the height and width viewed as collapsible spatial dimensions. The matrix $R$ means that given a point of the invisible regions, it gives the pair-wised relevance of the point with the points of all visible regions. Since the relationship matrix contains value for each point, a matrix multiplication of $R$ and the input feature can be applied to extract related information from the visible. At last, the outputs of the two streams are concatenated with the input feature $F_i^{in}$, and a $1\times1$ convolution layer is followed to reduce the channel number same with $F_i^{in}$.

The loss function to optimize our content recovery network is formulated as follows:
\begin{equation}
  \label{eq:rgb_loss}
  \begin{aligned}
  \mathcal{L}_{c}\: =\: &\beta_1\:(\mathbb{E}_{\hat{I}_{o}}[\mathrm{log}(1-D_c(\hat{I}_{o}))] + \mathbb{E}_{I_{o}}[\mathrm{log}\:D_c(I_{o})]) + \\
  &\beta_2\:\mathcal{L}_{\ell 1}(\hat{I}_{o},\: I_{o}) + \beta_3\:\mathcal{L}_{prec}(\hat{I}_{o},\: I_{o}) + \\
  &\beta_4\:\mathcal{L}_{style}(\hat{I}_{o},\: I_{o}), \\
  \end{aligned}
\end{equation}
where $\mathcal{L}_{style}(*)$ denotes the style loss proposed in ~\cite{partialconv_eccv18}.

\section{The Amodal Human Perception Dataset}       \label{dataset}

In this section, we describe how we collect human images and obtain their annotations with minimal manual effort. Our method adopts a straightforward pipeline and takes advantage of existing works on human segmentation. As a result, the proposed Amodal Human Perception dataset, namely AHP, contains unoccluded human images with masks. Therefore the occlusion cases can be synthesized by pasting occluders from other datasets onto humans. Moreover, some informative statistics are summarized to analyze our dataset.


  \begin{figure}[!t]
    \centering
    \includegraphics[width=1.0\linewidth]{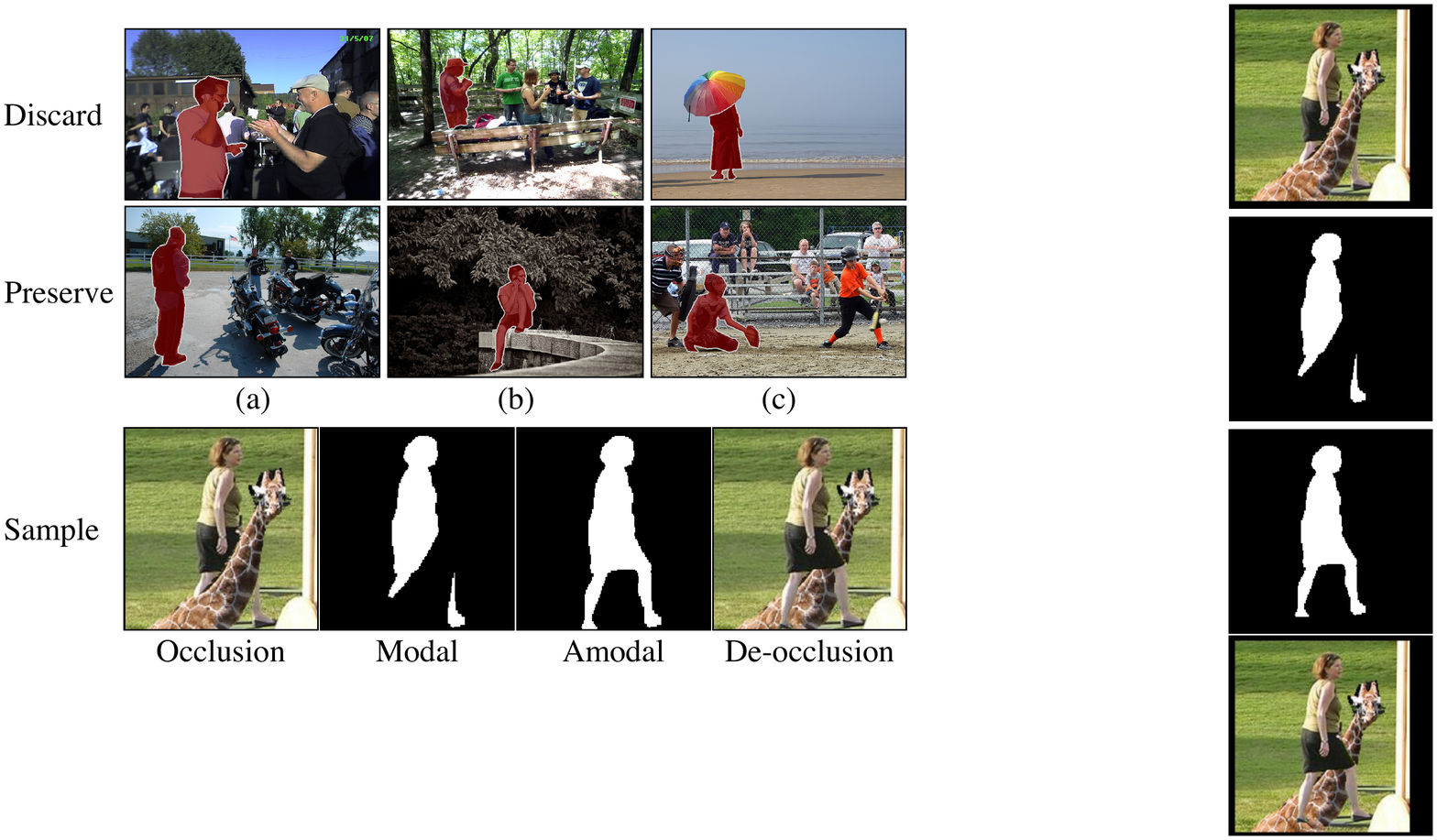}
    \caption{
    The first two rows show some `Discard' and `Preserve' exemplars and the last row demonstrates a synthesized occlusion image with ground-truths of the modal mask and the amodal mask and the final de-occlusion image.}
    \label{fig:discard_vs_preserve}
  \end{figure}

\subsection{Data Collection and Filtering}

Instead of collecting images of people from the Internet, we capitalize on several large instance segmentation and detection datasets to acquire human images. Then a segmentation model is applied to obtain masks for the humans with only box-level annotations. The reason to establish pixel-level annotations is that they can be utilized for occlusion synthesis. Finally, an image filtering scheme with manual effort is applied to construct our dataset. The detailed process is as follows.

\begin{figure*}[bthp]
  \centering
  \includegraphics[width=0.95\linewidth]{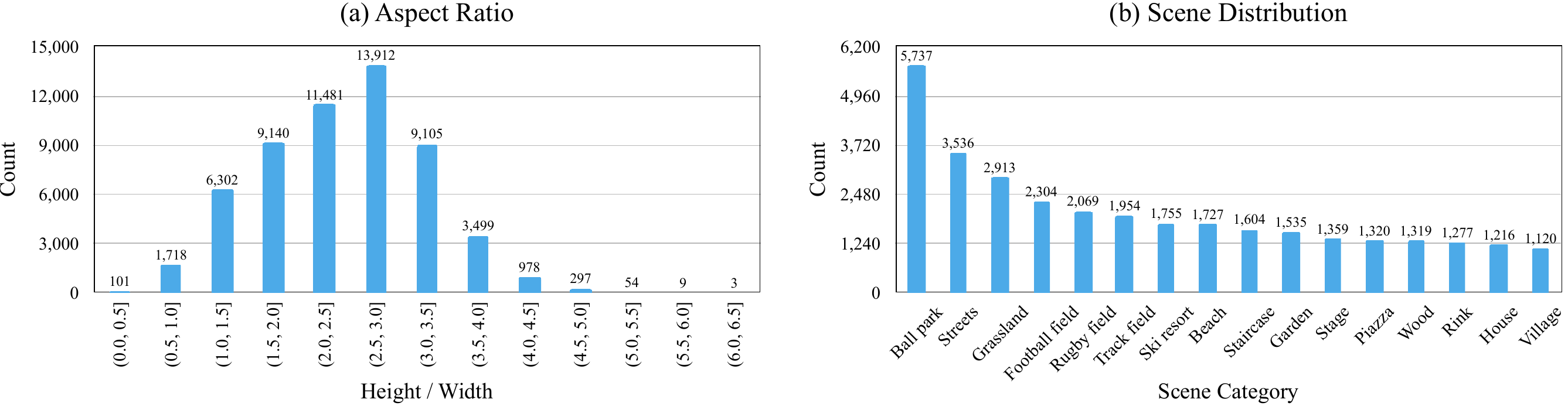}
  \caption{Chart (a) shows the distribution of aspect ratios and chart (b) demonstrates the distribution of scene categories in our dataset.}
  \label{fig:data_staticstics}
\end{figure*}

\begin{table}[b]
  \small
  \begin{center}
    \setlength{\tabcolsep}{2.6mm}     
    \begin{tabular}{l|ccc} 
    \toprule
                                    & COCO\dag    & OpenImages      & Objects365 \\
    \midrule
    Human Count          & $24,806$     & $314,359$      & $669,166$ \\
    Preserved Count     & $5,940$       & $17,677$         & $32,982$ \\
    \midrule
    Preserved Ratio      & $22.1$\%       & $5.62$\%          & $4.93$\% \\
    \bottomrule
    \end{tabular}
    \caption{Statistics of the human and the preserved counts. {COCO\dag} means COCO unites other datasets, \eg VOC.}
    \label{table:filtering_ratio}
  \end{center}
\end{table}

\noindent\textbf {(1) Image Acquisition.} We collect human images from several large-scale instance segmentation and detection datasets, including COCO~\cite{coco_eccv14}, VOC~\cite{voc_ijcv10} (with SBD~\cite{sbd_iccv11}), LIP~\cite{lip_cvpr17}, Objects365~\cite{objects365_iccv19} and OpenImages~\cite{openimages_arxiv18}. Since each instance in these datasets has been annotated with category label, we simply keep the instances labeled with `human' for further processing. In addition, human instances that are too small (\eg $< 300$ pixels) or have overlaps within the `gutter' (\eg $5$ pixels) along four image boundaries are dropped because parts of them are very likely to be out of view.

\noindent\textbf {(2) Human Segmentation.} The two largest datasets of Objects365 and OpenImages provide only box-level annotations, which means a great many human instances lack of pixel-level segmentation masks. Hence, a human segmentation model, \ie Deeplab~\cite{deeplabv3plus_eccv18} trained on private massive human data, is applied to obtain high-quality segmentation results. In specific, the box for each instance is enlarged by $3$ times in height and width to get a cropped image patch which will be fed into the model. Note we only keep segmentation results inside the original annotated box. Although the segmentation algorithm may fail on complicated cases, the issue will be addressed next. 

\noindent\textbf {(3) Filtering Scheme.} Since we aim at collecting unoccluded human instances with accurate masks, there is a need to retain satisfied samples. Three options are set up to route each sample: \textbf{Discard}: the human is occluded by other instances (\eg desk, car or other humans) or parts of him/her are out of view; \textbf{Preserve}: the human is not occluded and the segmentation is fine; \textbf{Refine}: the human is not occluded but the segmentation result is not satisfied. Some `Discard' and `Preserve' exemplars are demonstrated in Fig~\ref{fig:discard_vs_preserve}. Note for the case of `Preserve-(c)' that a baseball player wears a glove, we will keep it as the object affects the body structure not too much. But `Discard-(c)' is an opposite case. There are $7$ well-trained annotators filtering the collected human instances. Finally, the number of the `Preserve' samples reaches over $50,000$ and we think it is already enough to do meaningful research. The samples with the `Refine' label will be also released for potential exploration.

Once our dataset is built, occlusion training samples can be synthesized. As shown in the last row of Fig~\ref{fig:discard_vs_preserve}, an giraffe is cut out from other datasets and pasted onto the woman to form an occlusion case. Since we have the mask annotation of the woman (amodal mask), the ground-truths of the modal mask and the invisible content can be easily inferred.

\subsection{Data Statistics}

The AHP dataset consists of $56,599$ images following the aforementioned steps. Each sample costs about $3.26$ seconds on average. Table~\ref{table:filtering_ratio} shows the total number of the human category and the preserved ratio from each original dataset. It is noticeable that the preserved ratios of OpenImages and Objects365 are significantly smaller than COCO. Besides, the distributions of aspect ratios (height / width) and scene categories are shown in Fig~\ref{fig:data_staticstics}. Since the preserved humans are usually standing, it is reasonable that there are nearly half of the samples whose aspect ratios are between $1.5$ and $3.5$. To estimate the distribution of scene categories, a scene recognition model~\cite{inceptionresnet_arxiv16} is applied. Here we only show categories that contain more than $1,000$ samples.

\section{Experiments}   \label{experiments}

\subsection{Implementation Details} \label{impl_details}

\textbf{Networks and optimization.} The input sizes of the mask completion and the content recovery networks are both $256\times 256$. We adopt CenterMask~\cite{centermask_cvpr20} as our pretrained instance segmentation model $N_i$. HRNet~\cite{hrnet_cvpr19} is used to generate the pseudo labels of body parts. The hourglass modules of $N_m^{hg}$ and $N_a^{hg}$ and the Unet $N_c$ are inherited from \cite{pcnets_cvpr20} for fair comparison, and they are initialized with random weights. Both $D_m$ and $D_c$ adopt Patch-GAN~\cite{patchgan_cvpr17} and share a same structure with four convolution layers. We set $\lambda_1=\lambda_2=1, \lambda_3=0.1$ and $\beta_1=0.1, \beta_2=\beta_3=1,\beta_4=40$ in all experiments. We use SGD with momentum (batch $32$, lr $1\mathrm{e}{-3}$, iterations $48$k) and Adam~\cite{adam_arxiv14} (batch $16$, lr $1\mathrm{e}{-4}$, iterations $230$k) to optimize the completion and the recovery networks respectively. We use PyTorch~\cite{pytorch_nipsw17} framework with a NVIDIA Tesla V100 to conduct experiments.

\textbf{Data splits.} One advantage of our AHP dataset is that the distribution of human occlusion ratios can be controlled manually. Based on our observation, we set the probability density of occlusion ratios to $P_{0\sim 0.1}=P_{0.1\sim 0.2}=P_{0.3\sim 0.4}=1/3$ when training. We draw random instances from COCO~\cite{coco_eccv14} to synthesize occlusion cases. At present, our method adopts the simplest synthesis strategy and we leave the problem of how to generate better samples for future research. Our validation set contains $891$ images, which is augmented from $297$ integrated human samples each with three different occluders. The probability density of occlusion ratios for the validation split is set to $P_{0\sim 0.1}= P_{0.1\sim 0.2}=P_{0.3\sim 0.4}=P_{0.4\sim 0.5}=1/4$ and this set is fixed after generated. To verify the effectiveness of our method in real scenes, several photo editors collect a number of images and they move the foreground instances with similar depths onto the humans manually. Then these artificial occlusion cases are voted by another group of humans and only all passed samples are saved as our test set. The test set contains $56$ images due to its steep cost.

\textbf{Evaluation metrics.} To evaluate the quality of the completed masks, we adopt $\ell_1$ distance and Intersection over Union (IoU) as our metrics. For the recovered images, we adopt $\ell_1$ distance and Fréchet Inception Distance (FID)~\cite{fid_nips17} score which measures the similarity between the ground-truth images and our generated results. 




\subsection{Results}

\begin{table}[!t]
  \small
  \centering
  \setlength{\tabcolsep}{1.1mm}     
  \begin{tabular}{l|cc|cc}
  \toprule
  \multirow{2}*{Method} & \multicolumn{2}{c|}{Syn.}  & \multicolumn{2}{c}{Real} \\ 
  \cmidrule{2-5}
  ~                                      & $\ell_1\downarrow$ & IoU $\uparrow$ & $\ell_1\downarrow$ & IoU $\uparrow$ \\
  \midrule
  Mask-RCNN~\cite{maskrcnn_iccv17}       & $0.2402$           & $78.4/26.9$   & $0.2511$       & $75.6/23.8$ \\
  Deeplab~\cite{deeplabv3plus_eccv18}    & $0.2087$           & $70.7/20.9$   & $0.2179$       & $75.7/23.5$ \\
  Pix2Pix~\cite{pix2pix_cvpr17}          & $0.2329$           & $69.6/19.2$   & $0.2376$       & $68.0/16.0$ \\
  \midrule
  SeGAN~\cite{segan_cvpr18}              & $0.2545$           & $76.7/23.6$   & $0.2544$       & $77.7/19.0$ \\
  OVSR~\cite{caramodal_iccv19}           & $0.1830$           & $80.2/28.1$   & $0.1809$       & $82.9/25.6$ \\
  PCNets~\cite{pcnets_cvpr20}            & $0.1959$           & $83.1/29.1$   & $0.2218$       & $81.3/31.2$ \\
  \midrule
  Ours                                   & $\mathbf{0.1500}$  & $\mathbf{84.6/43.7}$ & $\mathbf{0.1635}$ & $\mathbf{86.1/40.3}$ \\
  \bottomrule
  \end{tabular}
  \caption{The comparison results of mask completion task on our AHP dataset. `Syn.' and `Real' denote synthesized and real validation images. Our method improves over other techniques both in $\ell_1$ error of the amodal masks and IoUs of the amodal and the invisible masks.}
  \label{table:maskmethods_vs}
  \end{table}

\textbf{Quantitative comparison.} We compare our method to recent state-of-the-art techniques on our AHP dataset. For the mask completion, some general methods widely applied in other fields like Mask-RCNN~\cite{maskrcnn_iccv17}, Deeplab~\cite{deeplabv3plus_eccv18} and Pix2Pix~\cite{pix2pix_cvpr17} are adopted. There are another group of methods specializing in handling the amodal perception task like SeGAN~\cite{segan_cvpr18}, OVSR~\cite{caramodal_iccv19} and PCNets~\cite{pcnets_cvpr20}. Among them, OVSR and PCNets declare they have achieved or surpassed state-of-the-art performance. As shown in Table~\ref{table:maskmethods_vs}, our method has lower $\ell_1$ error and better IoU results on amodal masks both on synthesized and real validation images. It is worth mentioning that we have a significant superiority over the invisible masks.
For the task of content recovery, we also selected the two types of general and specialized methods to compare. As shown in Table~\ref{table:rgbmethods_vs}, our method again performs over others on both validation sets.

\begin{figure*}[!t]
    \centering
    \includegraphics[width=0.96\linewidth]{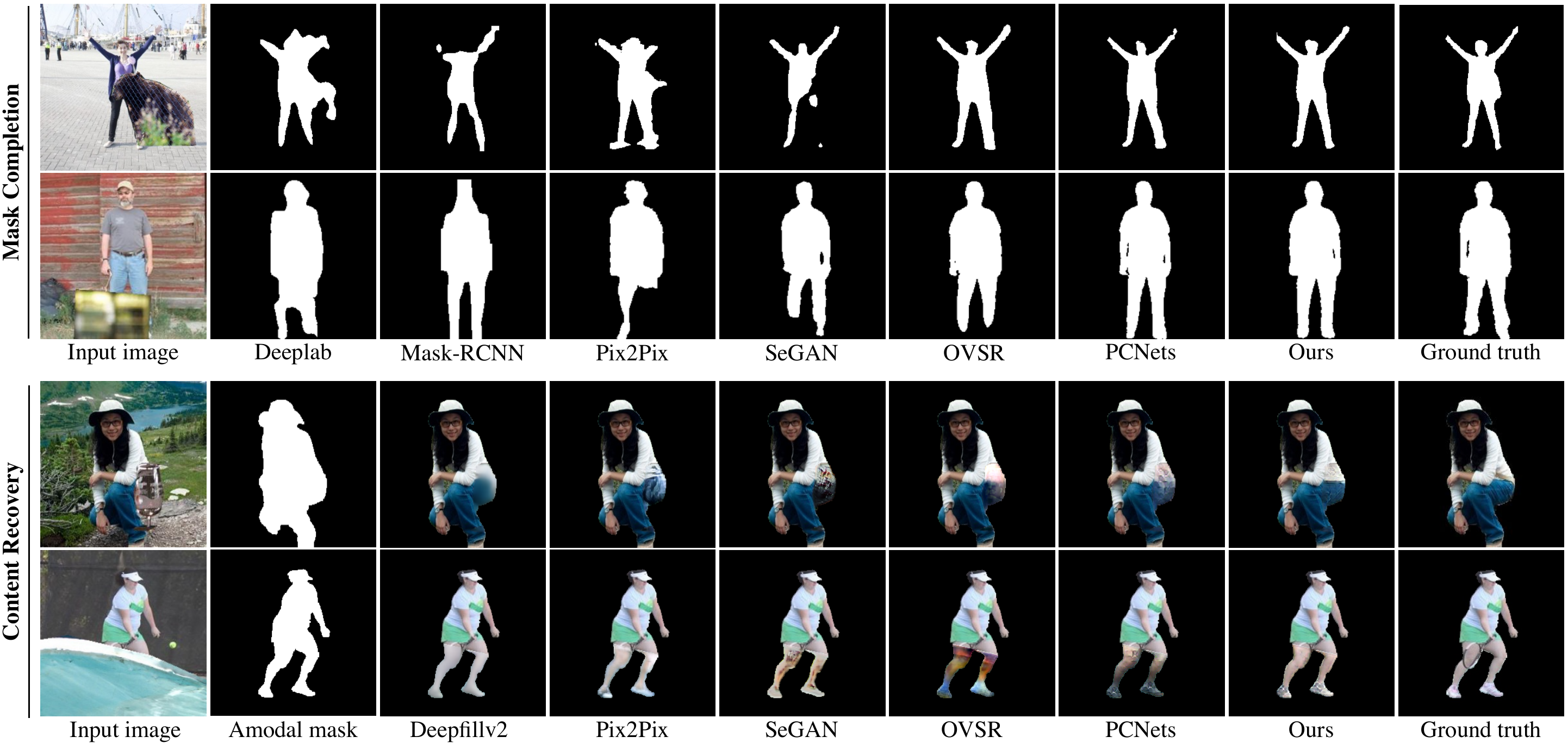}
    \caption{Qualitative comparison results of the mask completion and the content recovery tasks on our AHP dataset.}
    \label{fig:fig_res_vs}
  \end{figure*}

\textbf{Qualitative comparison.} Fig~\ref{fig:fig_res_vs} shows a sample of predicted amodal masks and recovered images on our AHP dataset. For the mask completion task, Pix2Pix~\cite{pix2pix_cvpr17} and SeGAN~\cite{segan_cvpr18} have difficulties at completing the occluded humans and their results are not satisfying. Our method has more reasonable and better predicted masks compared to OVSR~\cite{caramodal_iccv19} and PCNets~\cite{pcnets_cvpr20}. For the task of content recovery, Deepfillv2~\cite{gatedconv_iccv19} seems blurring and there are apparent artifacts in the results of Pix2Pix~\cite{pix2pix_cvpr17} and SeGAN~\cite{segan_cvpr18}. Again our method has better recovery performance compared to OVSR~\cite{caramodal_iccv19} and PCNets~\cite{pcnets_cvpr20}. More results are provided in the supplementary material.

\begin{table}[!t]
  \small
  \centering
  \setlength{\tabcolsep}{2.5mm}     
  \begin{tabular}{l|cc|cc}
  \toprule
  \multirow{2}*{Method} & \multicolumn{2}{c|}{Syn.}  & \multicolumn{2}{c}{Real} \\ 
  \cmidrule{2-5}
  ~                                   & $\ell_1\downarrow$ & FID $\downarrow$ & $\ell_1\downarrow$ & FID $\downarrow$ \\
  \midrule
  Pix2Pix~\cite{pix2pix_cvpr17}       & $0.1126$           & $19.66$          & $0.1031$ & $29.63$ \\
  Deepfillv2~\cite{gatedconv_iccv19}  & $0.1127$           & $21.61$          & $0.1026$ & $32.48$ \\
  \midrule
  SeGAN~\cite{segan_cvpr18}           & $0.1122$           & $23.01$          & $0.1027$ & $35.21$ \\
  OVSR~\cite{caramodal_iccv19}        & $0.0940$           & $27.15$          & $0.0917$ & $36.23$ \\
  PCNets~\cite{pcnets_cvpr20}         & $0.0936$           & $18.50$          & $0.0911$ & $28.30$ \\
  \midrule
  Ours                                & $\mathbf{0.0519}$  & $\mathbf{13.85}$ & $\mathbf{0.0617}$ & $\mathbf{19.49}$ \\
  \bottomrule
  \end{tabular}
  \caption{The comparison results of content recovery task on our AHP dataset. `Syn.' and `Real' denote synthesized and real validation images.}
  \label{table:rgbmethods_vs}
  \end{table}

\subsection{Ablation Study} \label{ab_study}

To understand our framework further, we conduct extensive experiments on the synthesized and fixed validation images (Sec~\ref{impl_details}) to prove the effectiveness of our method.

\begin{table}[!b]
  \small
  \begin{center}
    \setlength{\tabcolsep}{0.9mm}     
    \begin{tabular}{lcccc|c} 
    \toprule
                  & Discriminator   & Modal       & Parsing     & Templates  & IoU $\uparrow$ \\
        \midrule
        1        &               &             &                    &             & $77.5/84.0/30.3$ \\
        2        & \checkmark    &             &                    &             & $77.5/84.3/30.0$ \\
        3        & \checkmark    &  \checkmark &                    &             & $81.7/83.1/34.6$ \\
        \midrule
        4        & \checkmark    &  \checkmark & \checkmark         &             & $84.0/83.9/40.0$ \\
        5        & \checkmark    &  \checkmark &                    & \checkmark  & $82.4/83.3/38.5$ \\
        \midrule
        6        & \checkmark    &  \checkmark & \checkmark  & \checkmark & $\mathbf{84.7}/\mathbf{84.6}/\mathbf{43.7}$ \\
    \bottomrule
    \end{tabular}
    \caption{Ablation study of the refined mask completion network. The three columns of each IoU result represent the modal, the amodal and the invisible masks respectively.}
    \label{table:abs_pcnetm}
  \end{center}
\end{table}

\textbf{Refined mask completion network.} Table~\ref{table:abs_pcnetm} shows the results. The baseline takes the input image $I_s$ and the initial mask $M_i$, then outputs the amodal mask $\hat{M}_a$ by networks $N_m^{hg}$ and $N_a^{hg}$ (line 1). The discriminator $D_m$ improves the quality of the amodal mask $M_a$ by $0.3\%$ (line 2).
To obtain precise invisible mask, the introduction of the modal segmentation $M_m$ significantly refines the modal mask result by $4.2\%$ but degrades the performance of the amodal by $1.2\%$. Fortunately the final invisible mask has $4.3\%$ improvements (line 3). It shows that the modal segmentation has negative effects on the amodal completion, and we speculate obtaining the amodal mask requires analogous feature both in the occluded and visible portions of the humans.
To solve the problem, the accompanying parsing branches ($M_m^p,M_a^p$) are leveraged to bring in extra semantic guidance and template pose masks are utilized. The parsing improves $2.3\%$ and $0.8\%$ for the modal and amodal tasks respectively (line 4) and the network benefits from the templates by $0.7\%$ and $0.2\%$ (line 5). It is noticeable that the IoU metric of the invisible mask boosts $5.4\%$ and $3.9\%$ of the two proposed modules.
Finally, after we unite these parts together, the IoUs of the modal and invisible masks are remarkably improved by $7.2\%$ and $13.4\%$ respectively compared to the baseline (line 6 vs. line 1).

\begin{table}[!t]
  \small
  \begin{center}
    \setlength{\tabcolsep}{1.3mm}     
    \begin{tabular}{c|cccccccc} 
    \toprule
    $w$              & $0.0$  & $0.1$    & $0.3$   & $0.5$   & $0.7$   & $0.9$   & $1.0$\\
    \midrule
    FID $\downarrow$ & $18.04$ & $18.19$ & $\mathbf{17.85}$ & $18.74$ & $18.61$ & $19.11$ & $19.71$\\
    \bottomrule
    \end{tabular}
    \caption{Ablation study of the proportion of the background.}
    \label{table:abs_fgbg}
  \end{center}
\end{table}

\textbf{Parsing guided content recovery network.} In Table~\ref{table:abs_fgbg}, we analyze the effect of the background first. Intuitively the network should interpolate the invisible portions referring to the visible parts of the human only. However, the background may support context information to aid where the content can be imitated. Therefore, the new input image can be written as: $I_s^{'} = I_s*M_a+I_s*(1-M_a)*w$, where $w$ is the proportion of the background. 
The baseline is Unet with partial convolution as ~\cite{pcnets_cvpr20} with $w=1$.
The experiments show that it gains $1.9$ points when $w=0.3$. 
Further, our proposed PGA module is disassembled to two attention streams and analyzed as shown in Table~\ref{table:abs_pcnetc}. The two streams are evaluated individually and the comparison with simply cascading the two streams shows our structure has better performance. Specifically, the first attention stream (Attention.B) separating different body parts boosts the performance by $2.94$ points and the second stream (Attention.T) of establishing the relationship between the visible context and the invisible regions gains $2.3$ points. 
Assembling the two streams in a cascade manner yields $5.04$ points improvement. Lastly, our proposed structure depicted on the right of Fig~\ref{fig:rgbcomp_net} further boosts extra $0.8$ points.

\begin{table}[!t]
  \small
  \begin{center}
    \setlength{\tabcolsep}{1.6mm}     
    
    \begin{tabular}{lcccc|c} 
    \toprule
                  & Bg ($0.3$) & Attention.B      & Attention.T       & Structure   & FID $\downarrow$ \\
    \midrule
        1         &            &            &             &          & $19.66$    \\                  
        2         & \checkmark &            &             &          & $17.85$  \\
        3         &            & \checkmark &             &          & $16.76$    \\
        4         &            &            &  \checkmark &          & $17.37$  \\
    \midrule
        5         & \checkmark & \checkmark &  \checkmark & Cascade  & $14.66$  \\
        6         & \checkmark & \checkmark &  \checkmark & Fusion      & $\mathbf{13.85}$  \\

    \bottomrule
    \end{tabular}
    \caption{Ablation study of the parsing guided content recovery network. `Bg' denotes the background, and the columns `Attention.B' and `Attention.T' correspond to the two attention streams.There are two structures to assemble them: `Cascade' and `Fusion'.}
    \label{table:abs_pcnetc}
  \end{center}
\end{table}

\section{Conclusion}

In this paper, we tackle \textit{human de-occlusion} which is a more special and important task compared with de-occluding general objects. By refining the initial masks from the segmentation model and completing the modal masks, our network is able to precisely predict the invisible regions. Then the content recovery network equipped with our proposed PGA module recovers the appearance details. Our AHP dataset has prominent advantages compared to the current amodal perception datasets. Extended studies on how to generate more realistic samples and aggregate deep features of the two tasks will be explored.

\section{Acknowledgements}
This work was in part supported by National Natural Science Foundation of China (No. 61876212) and Zhejiang Lab (No. 2019NB0AB02).

\clearpage
{\small
    \bibliographystyle{ieee_fullname}
    \bibliography{reference}
}

\end{document}